\pdfoutput=1

\documentclass[11pt]{article}

\usepackage{acl}

\usepackage{times}
\usepackage{latexsym}
\usepackage{booktabs}
\usepackage{tabularx}

\usepackage[T1]{fontenc}

\usepackage[utf8]{inputenc}

\usepackage{microtype}
\usepackage{caption}
\usepackage{subcaption}
\usepackage{adjustbox}
\usepackage{multirow}
\usepackage{soul}
\usepackage{amsmath}

\title{Answering Open-Domain Multi-Answer Questions via \\
a Recall-then-Verify Framework}

\author{
    Zhihong Shao, Minlie Huang\thanks{*Corresponding author: Minlie Huang.}\\
    The CoAI group, DCST, Tsinghua University, Institute for Artificial Intelligence; \\
    State Key Lab of Intelligent Technology and Systems; \\
    Beijing National Research Center for Information Science and Technology; \\
    Tsinghua University, Beijing 100084, China \\
    {\tt szh19@mails.tsinghua.edu.cn, aihuang@tsinghua.edu.cn}}

\begin{document}
\maketitle
\begin{abstract}
Open-domain questions are likely to be open-ended and ambiguous, leading to multiple valid answers. Existing approaches typically adopt the \textit{rerank-then-read} framework, where a reader reads top-ranking evidence to predict answers.
According to our empirical analysis, this framework faces three problems: \textit{first}, to leverage a large reader under a memory constraint, the reranker should select only a few relevant passages to cover diverse answers, while balancing relevance and diversity is non-trivial;
\textit{second}, the small reading budget prevents the reader from accessing valuable retrieved evidence filtered out by the reranker; \textit{third}, when using a generative reader to predict answers all at once based on all selected evidence, whether a valid answer will be predicted also pathologically depends on the evidence of some other valid answer(s).
To address these issues, we propose to answer open-domain multi-answer questions with a \textit{recall-then-verify} framework, which separates the reasoning process of each answer so that we can make better use of retrieved evidence while also leveraging large models under the same memory constraint.
Our framework achieves state-of-the-art results on two multi-answer datasets, and predicts significantly more gold answers than a \textit{rerank-then-read} system that uses an oracle reranker.
\end{abstract}

\section{Introduction}
Open-domain question answering \cite{DBLP:conf/trec/Voorhees99,DBLP:conf/acl/ChenFWB17} is a long-standing task where a question answering system goes through a large-scale corpus to answer information-seeking questions.
Previous work typically assumes that there is only one well-defined answer for each question, or only requires systems to predict one correct answer, which largely simplifies the task.
However, humans may lack sufficient knowledge or patience to frame very specific information-seeking questions, leading to open-ended and ambiguous questions with multiple valid answers. According to \citet{DBLP:conf/emnlp/MinMHZ20}, over 50\% of a sampled set of Google search queries \cite{DBLP:journals/tacl/KwiatkowskiPRCP19} are ambiguous.
Figure \ref{tab:ambigqa_sample} shows an example with at least three interpretations.
As can be seen from this example, the number of valid answers depends on both questions and relevant evidence, which challenges the ability of comprehensive exploitation of evidence from a large-scale corpus.
\begin{table}[!t]
    \centering
    \small
    \begin{tabularx}{0.49\textwidth}{X}
        \hline
        \textbf{Original Question:} When did \textcolor{blue}{[You Don't Know Jack]} come out? \\
        \hline
        \textbf{Interpretation \#1:} When did the first video game called \textcolor{blue}{[You Don't Know Jack]} come out?\\
        \textbf{Evidence \#1:} You Don't Know Jack is a video game released in 1995, and the first release in ... \\
        \textbf{Answer \#1:} 1995 \\
        \hline
        \textbf{Interpretation \#2:} When did the Facebook game \textcolor{blue}{[You Don't Know Jack]} come out on Facebook? \\
        \textbf{Evidence \#2:} In 2012, Jackbox Games developed and published a social version of the game on Facebook ... \\
        \textbf{Answer \#2:} 2012 \\
        \hline
        \textbf{Interpretation \#3:} When did the film \textcolor{blue}{[You Don't Know Jack]} come out? \\
        \textbf{Evidence \#3:} ``You Don't Know Jack'' premiered April 24, 2010 on HBO. \\
        \textbf{Answer \#3:} April 24, 2010 \\
        \hline
    \end{tabularx}
    \caption{An example of open-domain multi-answer questions. We display only a subset of valid answers. In fact, \textcolor{blue}{[You Don't Know Jack]} can also be a song.}
    \label{tab:ambigqa_sample}
\end{table}

Existing approaches mostly adopt the \textit{rerank-then-read} framework. A retriever retrieves hundreds or thousands of relevant passages which are later reranked by a reranker; a generative reader then predicts all answers in sequence conditioned on top-ranking passages.
With a fixed memory constraint\footnote{We follow \citet{DBLP:conf/emnlp/MinLCTH21} to constrain memory usage, which is usually a bottleneck of performance on open-domain question answering 
.}, there is a trade-off between the size of the reader and the number of passages the reader can process at a time.
According to \citet{DBLP:conf/emnlp/MinLCTH21}, provided that the reranker is capable of selecting a small set of highly-relevant passages with high coverage of diverse answers, adopting a larger reader can outperform a smaller reader using more passages. 
However, as shown by Section \ref{sec:analyses}, this framework
is faced with three problems: 
\textit{first}, due to the small reading budget, the reranker has to balance relevance and diversity, which is non-trivial as it is unknown beforehand that which answers should be distributed with more passages to convince the reader and which answers can be safely distributed with less to save the budget for the other answers;
\textit{second}, the reader has no access to more retrieved evidence that may be valuable but is filtered out by the reranker, while combining information from more passages was found to be beneficial to open-domain QA \cite{DBLP:conf/eacl/IzacardG21};
\textit{third}, as the reader predicts answers in sequence all at once,
the reader learns pathological dependencies among answers, i.e., whether a valid answer will be predicted also depends on passages that cover some other valid answer(s), while ideally, prediction of a particular answer should depend on the soundness of associated evidence itself.

To address these issues, we propose to answer open-domain multi-answer questions with a \textit{recall-then-verify} framework.
Specifically, we first use an answer recaller to predict possible answers from each retrieved passage individually; this can be done with high recall, even when using a weak model for the recaller, but at the cost of low precision due to insufficient evidence to support or refute a candidate. We then aggregate retrieved evidence relevant to each candidate, and verify each candidate with a large answer verifier. 
By separating the reasoning process of each answer, our framework avoids the problem of multiple answers sharing a limited reading budget, and makes better use of retrieved evidence while also leveraging strong large models under the same memory constraint.

Our contributions are summarized as follows:
\begin{itemize}
    \item We empirically analyze the problems faced by the \textit{rerank-then-read} framework when dealing with open-domain multi-answer QA.

    \item To address these issues, we propose to answer open-domain multi-answer questions with a \textit{recall-then-verify} framework, which makes better use of retrieved evidence while also leveraging the power of large models under the same memory constraint.

    \item Our framework establishes a new state-of-the-art record on two multi-answer QA datasets with significantly more valid predictions.
\end{itemize}

\section{Related Work}
Open-domain QA requires question answering systems to answer factoid questions by searching for evidence from a large-scale corpus such as Wikipedia \cite{DBLP:conf/trec/Voorhees99,DBLP:conf/acl/ChenFWB17}.
The presence of many benchmarks has greatly promoted the development of this community, such as questions from real users like NQ \cite{DBLP:journals/tacl/KwiatkowskiPRCP19} and W\textsc{eb}Q\textsc{uestions} \cite{DBLP:conf/emnlp/BerantCFL13}, and trivia questions like Quasar-T \cite{DBLP:journals/corr/DhingraMC17} and TriviaQA \cite{DBLP:conf/acl/JoshiCWZ17}. All these benchmarks either assume that each question has only one answer with several alternative surface forms, or only require a system to predict one valid answer.
A typical question answering system is a pipeline as follows: an efficient retriever retrieves relevant passages using sparse \cite{DBLP:conf/acl/MaoHLSG0C20,DBLP:conf/naacl/ZhaoLL21} or dense \cite{DBLP:conf/emnlp/KarpukhinOMLWEC20,DBLP:conf/iclr/XiongXLTLBAO21,DBLP:conf/iclr/IzacardG21,DBLP:journals/tacl/relevance-colbert} representations; an optional passage reranker \cite{DBLP:conf/sigir/AsadiL13,DBLP:journals/corr/abs-1901-04085,Nogueira_2020} further narrows down the evidence; an extractive or generative reader \cite{DBLP:conf/eacl/IzacardG21,DBLP:conf/acl/0002SLHCG20} predicts an answer conditioned on retrieved or top-ranking passages. Nearly all previous work focused on locating passages covering at least one answer, or tried to predict one answer precisely.

However, both \citet{DBLP:journals/tacl/KwiatkowskiPRCP19} and \citet{DBLP:conf/emnlp/MinMHZ20} reported that there is genuine ambiguity in open-domain questions, resulting in multiple valid answers.
To study the challenge of finding all valid answers for open-domain questions, \citet{DBLP:conf/emnlp/MinMHZ20} proposed a new benchmark called A\textsc{mbig}QA where questions are annotated with as many answers as possible. In this new task, the passage reranker becomes more vital in the \textit{rerank-then-read} framework, particularly when only a few passages are allowed to feed a large reader due to memory constraints. This is because the reranker has to ensure that top-ranking passages are highly relevant and also cover diverse answers.
Despite state-of-the-art performance on A\textsc{mbig}QA \cite{DBLP:conf/emnlp/MinLCTH21}, according to our empirical analysis, applying the \textit{rerank-then-read} framework to open-domain multi-answer QA faces the following problems: balancing relevance and diversity is non-trivial for the reranker due to unknown effect on the performance of the subsequent reader; when using a large reader under a fixed memory constraint, the small reading budget prevents it from making use of more retrieved evidence that is valuable but filtered out; when using a generative reader to predict all answers in sequence based on all selected evidence,
it learns pathological dependencies among answers.
To address these issues, we propose to tackle this task with a \textit{recall-then-verify} framework, which separates the reasoning process of each answer with a higher level of evidence usage while also leveraging large models under the same memory constraint.

Some previous work argued that a reader can be confused by similar but spurious passages, resulting in wrong predictions. Therefore, they proposed answer rerankers \cite{DBLP:conf/iclr/WangY0ZGCWKTC18,DBLP:conf/acl/WuWLHWLLL18,DBLP:conf/aaai/HuWPHYL19,DBLP:conf/naacl/IyerMMY21} to rerank top predictions from readers.
Our framework is related to answer reranking but with two main differences. 
First, a reader typically aggregates available evidence and already does a decent job of answer prediction even without answer reranking; an answer reranker is introduced to filter out hard false positive predictions from the reader. By contrast, our answer recaller aims at finding possible answers with high recall, most of which are invalid. Evidence focused on each answer is then aggregated and reasoned about by our answer verifier. It is also possible to introduce another model analogous to an answer reranker to filter out false positive predictions from our answer verifier.
Second, answer reranking typically compares answer candidates to determine the most valid one, while our answer verifier selects multiple valid answers mainly based on the soundness of their respective evidence but without comparisons among answer candidates.

\section{Task Formulation}
Open-domain multi-answer QA can be formally defined as follows: given an open-ended question $q$, a question answering system is required to make use of evidence from a large-scale text corpus $\mathcal{C}$ and predict a set of valid answers $\{a_1, a_2, ..., a_n\}$. Questions and their corresponding answer sets are provided for training.


\noindent
\textbf{Evaluation}
To evaluate passage retrieval and reranking, we adopt the metric MR\textsc{ecall}@$k$ from \cite{DBLP:conf/emnlp/MinLCTH21}, which measures whether the top-$k$ passages cover at least $k$ distinct answers (or $n$ answers if the total number of answers $n$ is less than $k$). To evaluate question answering performance, we follow \cite{DBLP:conf/emnlp/MinMHZ20} to use F1 score between gold answers and predicted ones.

\section{Rerank-then-Read Framework}
In this section, we will briefly introduce the representative and state-of-the-art \textit{rerank-then-read} pipeline from \cite{DBLP:conf/emnlp/MinLCTH21} for open-domain multi-answer questions, and provide empirical analysis of this framework.

\subsection{Passage Retrieval}
Dense retrieval is widely adopted by open-domain question answering systems \cite{DBLP:conf/nips/MinBACC0GHLPRRK20}. A dense retriever measures relevance of a passage to a question by computing the dot product of their semantic vectors encoded by a passage encoder and a question encoder, respectively. Given a question, a set of the most relevant passages, denoted as $\mathcal{B}$ ($|\mathcal{B}| \ll |\mathcal{C}|$), is retrieved for subsequent processing.

\subsection{Passage Reranker}
To improve the quality of evidence, previous work \cite{Nogueira_2020,DBLP:conf/acl/GaoZNSWNZNAX20} finds it effective to utilize a passage reranker, which is more expressive than a passage retriever, to rerank retrieved passages, and select the $k$ best ones to feed a reader for answer generation ($k < |\mathcal{B}|$). With a fixed memory constraint, there is a trade-off between the number of selected passages and the size of the reader. As shown by \cite{DBLP:conf/emnlp/MinLCTH21}, with good reranking, using a larger reader is more beneficial. To balance relevance and diversity of evidence, \citet{DBLP:conf/emnlp/MinLCTH21} proposed a passage reranker called JPR for joint modeling of selected passages. Specifically, they utilized T5-base \cite{DBLP:journals/jmlr/RaffelSRLNMZLL20} to encode retrieved passages following \cite{DBLP:conf/eacl/IzacardG21} and decode the indices of selected passages autoregressively using a tree-decoding algorithm.
JPR is designed to seek for passages that cover new answers, while also having the flexibility to select more passages covering the same answer, especially when there are less than $k$ answers for the question.

\subsection{Reader}
A reader takes as input the top-ranking passages, and predicts answers. \citet{DBLP:conf/emnlp/MinLCTH21} adopted a generative encoder-decoder reader initialized with T5-3b, and used the fusion-in-decoder method from \cite{DBLP:conf/eacl/IzacardG21} which efficiently aggregates evidence from multiple passages. Specifically, each passage is concatenated with the question and is encoded independently by the encoder; the decoder then attends to the representations of all passages and generates all answers in sequence, separated by a \texttt{[SEP]} token.

\subsection{Empirical Analysis}
\label{sec:analyses}
To analyze performance of the \textit{rerank-then-read} framework for open-domain multi-answer questions, we built a system that resembles the state-of-the-art pipeline from \cite{DBLP:conf/emnlp/MinLCTH21} but with two differences\footnote{Code and models from \cite{DBLP:conf/emnlp/MinLCTH21} were not publicly available in the period of this work.\label{footnote:reason}}.
First, we used the retriever from \cite{DBLP:conf/iclr/IzacardG21}.
Second, instead of using JPR, we used an oracle passage reranker (OPR): a passage $p$ is ranked higher than another passage $p^\prime$ if and only if 1) $p$ covers some answer while $p^\prime$ covers none 2) or both $p$ and $p^\prime$ cover or fail to cover some answer but $p$ has a higher retrieval score.
Following \cite{DBLP:conf/emnlp/MinLCTH21}, we retrieved $|\mathcal{B}|$=100 Wikipedia passages, $k$=10 of which were selected by the reranker.
Table \ref{tab:oracle_rerank} shows model performance on a representative multi-answer dataset called A\textsc{mbig}QA \cite{DBLP:conf/emnlp/MinMHZ20}. Compared with JPR, OPR is better in terms of reranking, with similar question answering results\footnote{With the oracle knowledge of whether a passage contains a gold answer during reranking, OPR is probably still far from being a perfect reranker. Notably, we are not striving for a better \textit{rerank-then-read} pipeline for multi-answer questions, but use OPR as a representative case to analyze the problems a \textit{rerank-then-read} pipeline may face.}.

\begingroup
\setlength{\tabcolsep}{3pt} 
\renewcommand{\arraystretch}{1} 
\begin{table}[htp]
    \centering
	\adjustbox{max width=.35\textwidth}{
    \begin{tabular}{c|c|c|c}
        \hline
		\multirow{2}{*}{Model} & \multicolumn{2}{c|}{Reranking} & QA \\
		\cline{2-4}
		& MR\textsc{ecall}@$5$ & MR\textsc{ecall}@$10$ & F1 \\
        \hline
        JPR & 64.8/45.2 & 67.1/48.2 & \textbf{48.5}/\textbf{37.6} \\
        OPR & \textbf{67.7}/\textbf{46.5} & \textbf{70.3}/\textbf{51.2} & 48.4/37.0 \\
	    \hline
    \end{tabular}
    }
    \caption{Reranking results and Question Answering results on the dev set of A\textsc{mbig}QA using JPR and OPR. The two numbers in each cell are results on all questions and questions with multiple answers, respectively.}
    \label{tab:oracle_rerank}
\end{table}
\begingroup

Though 3,670 diverse gold answers are covered by OPR on the dev set, the reader predicts only 1,554 of them. Our empirical analysis and findings are detailed as follows.

\begin{figure}
     \centering
     \begin{subfigure}[b]{0.196\textwidth}
         \centering
         \includegraphics[width=\textwidth]{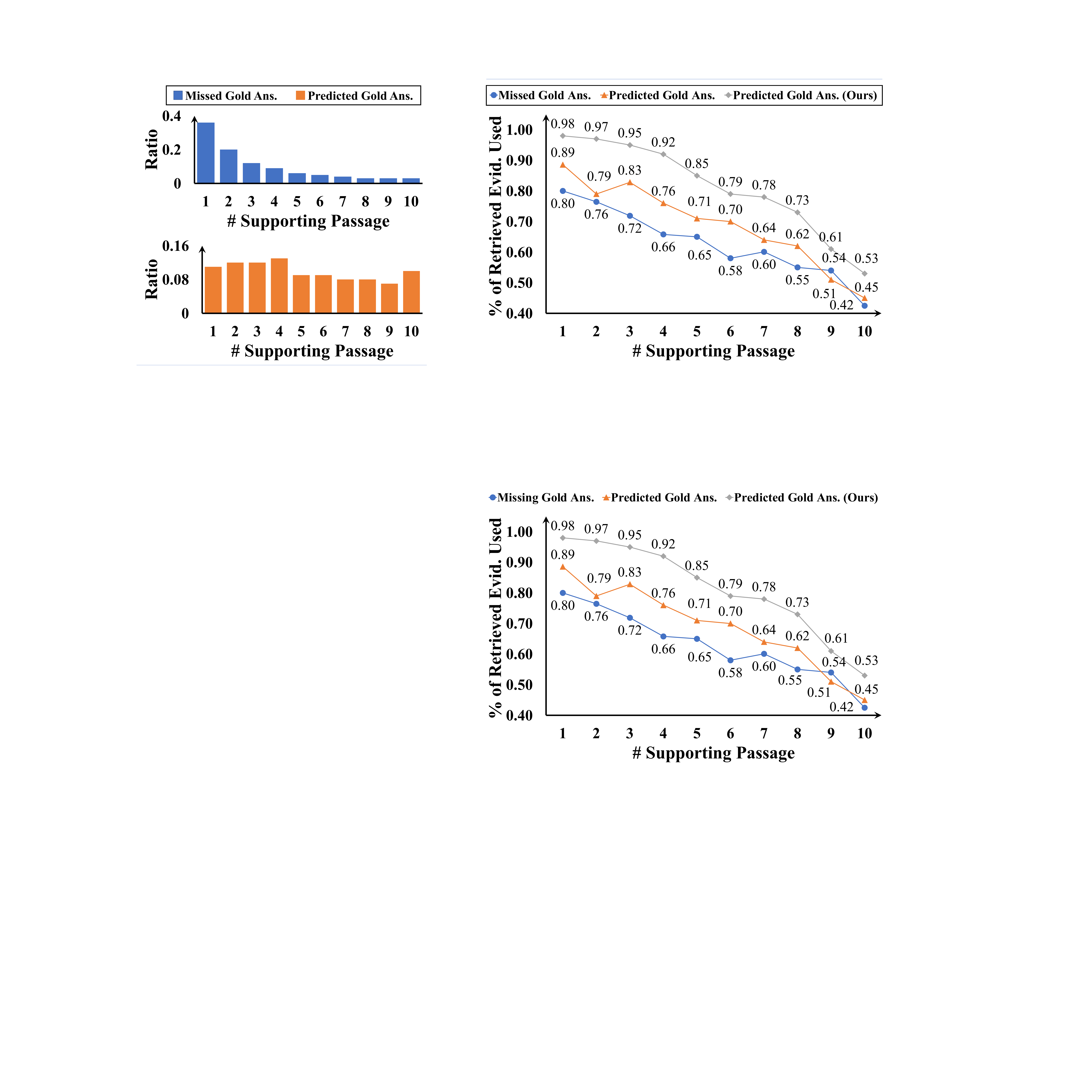}
         \caption{}
         \label{fig:missing_and_predcited_ans}
     \end{subfigure}
     \hfill
     \begin{subfigure}[b]{0.279\textwidth}
         \centering
         \includegraphics[width=\textwidth]{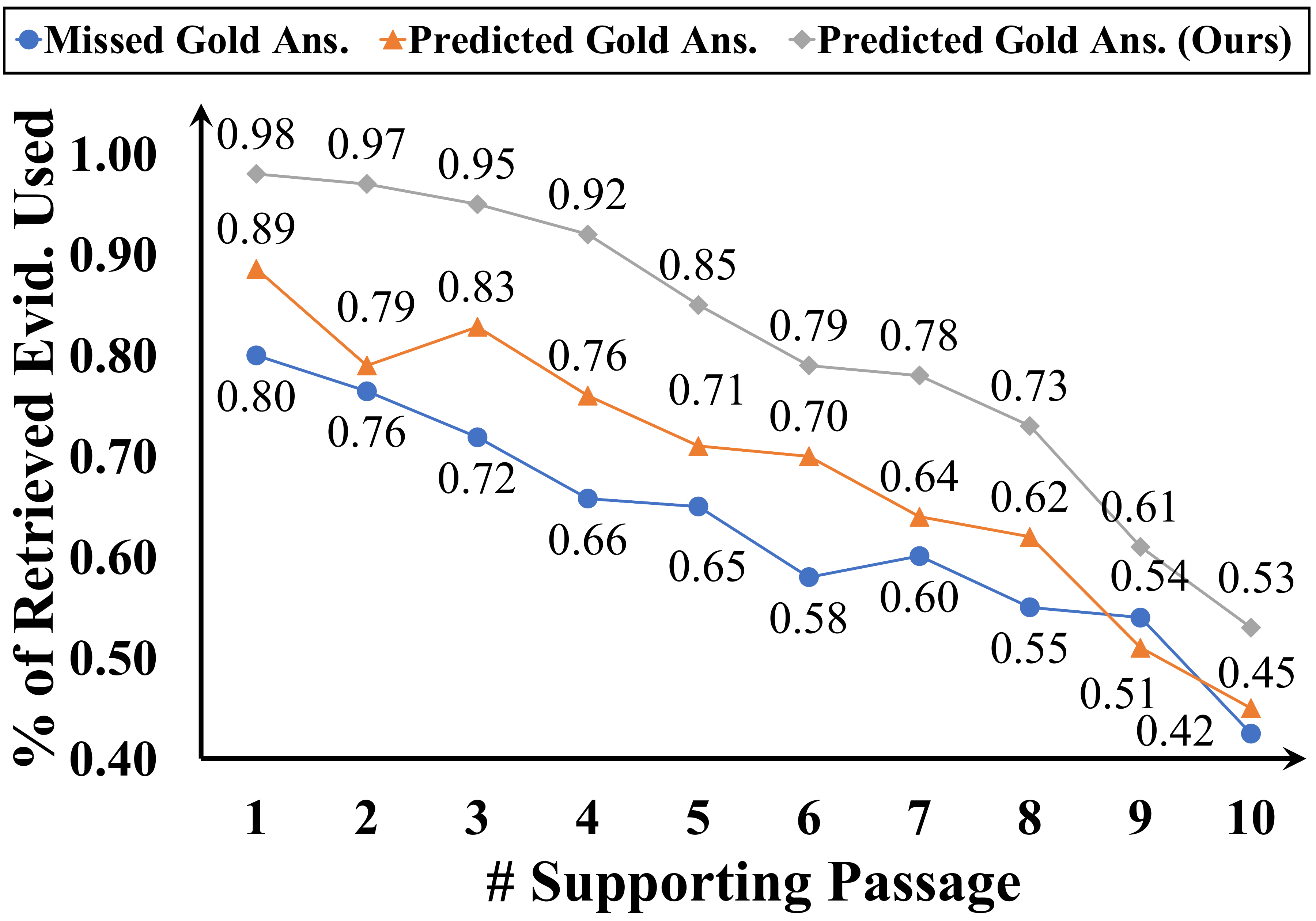}
         \caption{}
         \label{fig:evid_usage}
     \end{subfigure}
        \caption{
        Analysis of how well OPR (the reranker of a \textit{rerank-then-read} pipeline) balances relevance and diversity on questions with multiple answers in the dev set of A\textsc{mbig}QA.
        The number of retrieved passages is $|\mathcal{B}|$=100, and the number of passages selected by the reranker is $k$=10.
        Figure (a) shows the ratio of answers with different numbers of supporting passages selected, the \textcolor{blue}{top half} of which is for gold answers \textcolor{blue}{missed (top)} by the reader and the \textcolor{orange}{bottom half} is for \textcolor{orange}{predicted ones}.
        Figure (b) shows the ratio of retrieved supporting passages that are eventually used by the reader (or the verifier in our framework).
        }
        \label{fig:analyze_evid}
\end{figure}

\textit{(1) To leverage a large reader under a fixed memory constraint, a reranker should select only a few highly-relevant passages to cover diverse answers, while balancing relevance and diversity is non-trivial.}
As shown by Figure \ref{fig:missing_and_predcited_ans} (bottom), the number of selected supporting passages\footnote{We abuse the use of \textit{supporting passages} of an answer to refer to passages that cover the answer.} of predicted gold answers has a widespread distribution. There may be cases where redundant false positive evidence is selected and can be safely replaced with passages that cover other gold answers. However, it is non-trivial for the reranker to know beforehand whether a passage is redundant, and how many or which supporting passages of an answer are strong enough to convince the reader.

\textit{(2) Multiple answers sharing a small reading budget prevents a reader from using more evidence that may be valuable but is filtered out by the reranker.}
Due to the shared reading budget, it is inevitable that some answers are distributed with less supporting passages.
As shown by Figure \ref{fig:missing_and_predcited_ans}, a gold answer covered by OPR but missed by the reader generally has significantly less supporting passages fed to the reader (3.13 on average) than a predicted gold answer (5.08 on average), but not because of lacking available evidence. There is more evidence in retrieved passages for missed answers but filtered out by the reranker. As shown by Figure \ref{fig:evid_usage}, OPR has a much lower level of evidence usage for missed answers.

\begin{figure*}[!t]
    \centering
    \includegraphics[width=0.7\textwidth]{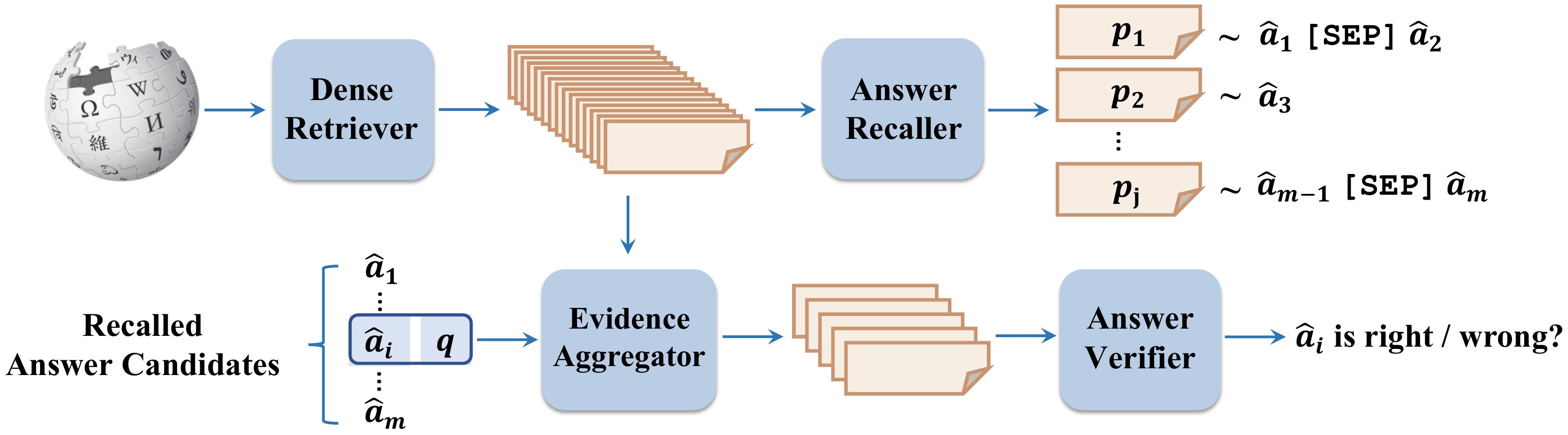}
    \caption{The \textit{recall-then-verify} framework we propose to answer open-domain multi-answer questions. We first use the answer recaller to guess possible answers with high recall, the evidence aggregator then aggregates retrieved evidence for each candidate, and finally, the answer verifier verifies each candidate based on its aggregated evidence. As the reasoning process of each answer is separated, and thanks to candidate-aware evidence aggregation, we can have a high level of evidence usage with a large verifier under a limited memory constraint.}
    \label{fig:framework}
\end{figure*}
\begin{figure}[!t]
    \centering
    \includegraphics[width=0.32\textwidth]{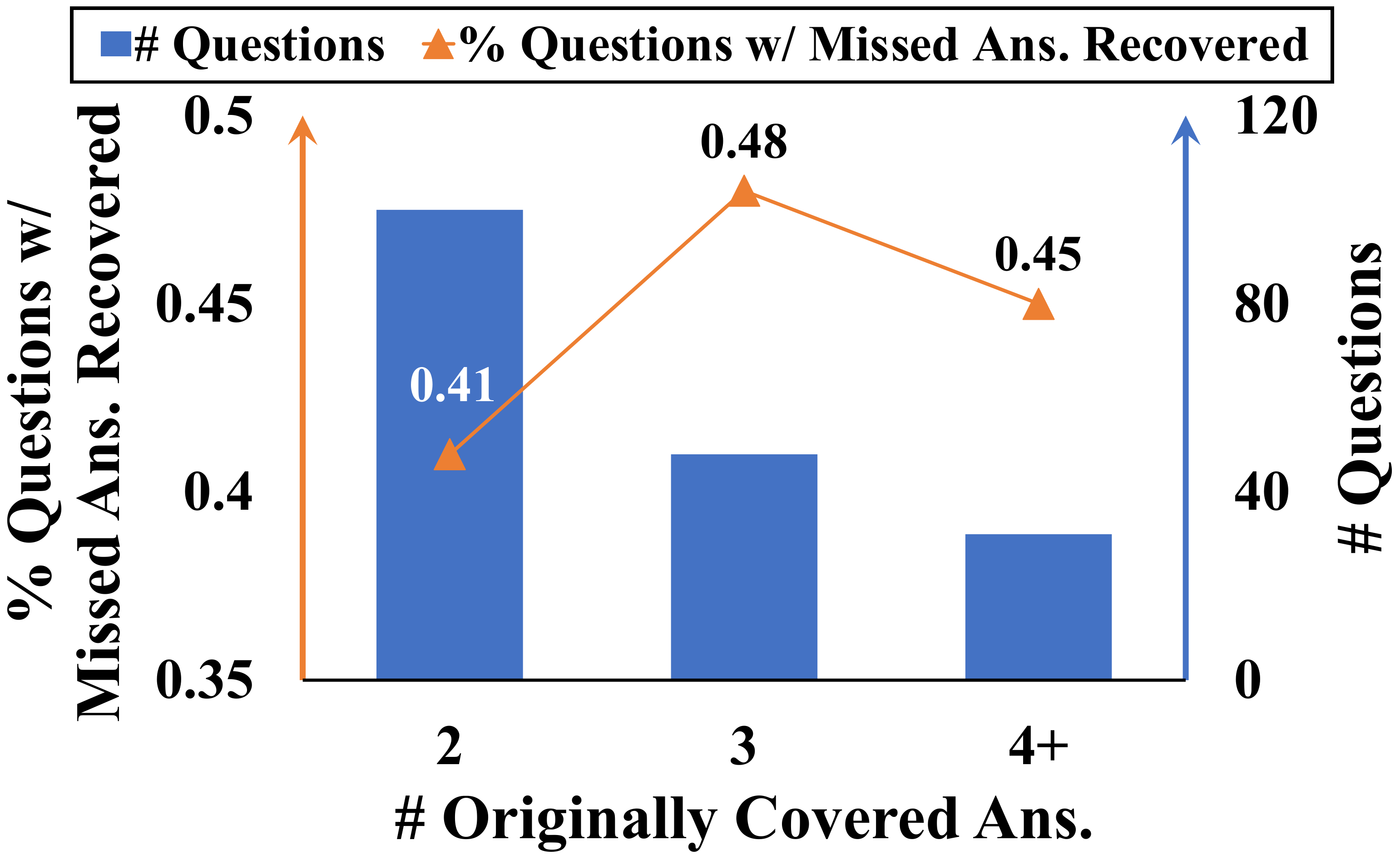}
    \caption{
    Analysis of the pathological dependencies among answers learned by the reader (of a \textit{rerank-then-read} pipeline) on the dev set of A\textsc{mbig}QA.
    The horizontal axis is the number of diverse answers covered by OPR.
    The left axis shows the ratio of questions for which the reader recovers some originally missed gold answer after adversarially removing the supporting passages of some originally predicted gold answer.}
    \label{fig:dep}
\end{figure}

\textit{(3) As the reader predicts answers all at once conditioned on all selected passages, whether a valid answer will be predicted also pathologically depends on evidence of some other valid answer(s), which partly accounted for the large number of gold answers missed by the reader.}
For verification, we attacked OPR's reader on the dev set of A\textsc{mbig}QA as follows: a question is a target if and only if 1) it has a gold answer covered by OPR but missed by the reader 2) and it has a predicted gold answer whose supporting passages cover no other gold answer; a successful attack on a targeted question means that a missed answer is recovered after removing a subset of supporting passages of some predicted answer\footnote{Removed passages were replaced with the same number of top-ranking passages that cover no gold answer, so that the number of passages fed to the reader remained unchanged.} without removing any supporting passage of the other gold answers.

There are 179 targeted questions; for 43.6\% of them, we successfully recovered at least one missed gold answer.
Figure \ref{fig:dep} shows the success rate breakdown on the number of answers covered by the reader's input, indicating that predictions tend to be brittle when the reader is fed with many diverse supporting passages.

One possible explanation of the pathological dependencies is that the reader implicitly compares the validity of answer candidates and predicts the most likely ones. However, for 40.0\% of successfully attacked questions, according to OPR, supporting passages of recovered missed answers are more relevant than those removed passages of predicted answers. Notably, \citet{DBLP:conf/emnlp/MinMHZ20} also had a similar observation on another \textit{rerank-then-read} pipeline, i.e., it is hard to argue that the predicted answers are more likely than the missed ones.

\section{Recall-then-Verify Framework}

\subsection{Overview}

To avoid the issues faced by the \textit{rerank-then-read} framework, we propose a \textit{recall-then-verify} framework, which separates the reasoning process of each answer so that answers (1) can be individually distributed with maximum supporting passages allowed on the same hardware (2) and are predicted mainly based on their own evidence.
Figure \ref{fig:framework} shows our framework.
Specifically, we first guess possible answers based on retrieved passages using an answer recaller, an evidence aggregator then aggregates evidence for each answer candidate, and finally, an answer verifier verifies each candidate and outputs valid ones.

\subsection{Answer Recaller}
Our answer recaller, based on T5, is trained to predict all gold answer(s) in sequence (separated by a \texttt{[SEP]} token) from each retrieved positive passage $p \in \mathcal{B}$ that cover some gold answer(s).
We also train the recaller to predict the ``irrelevant'' token given a negative passage so that the recaller can filter out negative candidates; the number of negatives per positive used for training is denoted as $\alpha_{neg}$.
The set of answer candidates recalled during inference is denoted as $\mathcal{A}=\{\hat{a}_1, \hat{a}_2, ..., \hat{a}_m\}$.
Though a passage may not contain strong enough evidence to support an answer, by exploiting semantic clues in the question and the passage (e.g., the answer type), it is sufficient for even a weak model to achieve high recall.
However, this is at the cost of low precision, which necessitates answer verification based on more supporting passages.

\subsection{Evidence Aggregator}
We aggregate evidence for each answer candidate from retrieved passages, which can be formulated as a reranking task, i.e., to rerank retrieved passages according to their relevance to a question-candidate pair, and select top-ranking ones for answer verification.
Our evidence aggregator resembles OPR: for a specific candidate $\hat{a}_i$, we encode the question-candidate pair with the retriever's question encoder; a passage $p$ is ranked higher than another passage $p^\prime$ if and only if 1) $p$ covers $\hat{a}_i$ while $p^\prime$ does not 2) or both $p$ and $p^\prime$ cover or fail to cover $\hat{a}_i$ but the semantic vector of $p$ is closer to that of the question-candidate pair.
We denote the top-$k$ relevant passages of $\hat{a}_i$ as $\mathcal{E}_i$.


\subsection{Answer Verifier}
Given a candidate $\hat{a}_i$ and its evidence $\mathcal{E}_i$, our answer verifier, based on T5-3b, predicts whether $\hat{a}_i$ is valid, using the fusion-in-decoder method from \cite{DBLP:conf/eacl/IzacardG21}.
Each passage from $\mathcal{E}_i$ is concatenated with the question and the candidate, and is encoded independently; the decoder then attends to the representations of all passages and is trained to produce the tokens ``right'' or ``wrong'' depending on whether the encoded candidate is valid or not\footnote{We have tried other verbalizers such as ``yes'' and ``no'', but found no significant difference.}.
During inference, we compute the validity score of a candidate by taking the normalized probability assigned to the token ``right'':
\begin{equation}
    \small
    \begin{split}
        P(&a_i\ \mbox{is valid}) =\\
        &\frac{\exp(\mbox{logit}(\mbox{``right''}|q, \hat{a}_i, \mathcal{E}_i))}{\sum_{t \in \{\mbox{``right''}, \mbox{``wrong''}\}}\exp(\mbox{logit}(t|q, \hat{a}_i, \mathcal{E}_i))}
    \end{split}
\end{equation}
Candidates with their validity scores higher than a threshold $\tau$ will be produced as final predictions.

\section{Experiments}
\subsection{Datasets}
We conducted experiments on two multi-answer QA datasets, whose statistics are shown in Table \ref{tab:data_stat}.

\noindent
\textbf{W\textsc{eb}QSP} \cite{DBLP:conf/acl/YihRMCS16} is a semantic parsing dataset for knowledge base question answering, where answers are a set of entities in Freebase. Following \cite{DBLP:conf/emnlp/MinLCTH21}, we repurposed this dataset for textual QA based on Wikipedia\footnote{\label{footnote:diff_split}Our train/dev split on W\textsc{eb}QSP is different from \citet{DBLP:conf/emnlp/MinLCTH21}'s, as their split was not publicly available in the period of this work.}.

\noindent
\textbf{A\textsc{mbig}QA} \cite{DBLP:conf/emnlp/MinMHZ20} originates from NQ \cite{DBLP:journals/tacl/KwiatkowskiPRCP19}, where questions are annotated with equally valid answers from Wikipedia.

\begingroup
\setlength{\tabcolsep}{3pt} 
\renewcommand{\arraystretch}{1} 
\begin{table}[htb]
    \centering
	\adjustbox{max width=.35\textwidth}{
    \begin{tabular}{cccccc}
        \hline
		\multirow{2}{*}{Dataset} & \multicolumn{3}{c}{\# Question} & \multicolumn{2}{c}{\# Answer} \\
		\cmidrule(lr){2-4} \cmidrule(lr){5-6}
        & Train & Dev & Test & Avg. & Median \\
        \hline
        W\textsc{eb}QSP & 2,752 & 245 & 1582 & 22.6 & 1.0 \\
        A\textsc{mbig}QA & 10,036 & 2,002 & 2,004 & 2.2 & 2.0 \\
	    \hline
    \end{tabular}
    }
    \caption{Statistics of multi-answer QA datasets. Statistics of answers are computed on the dev sets.}
    \label{tab:data_stat}
\end{table}
\begingroup

\subsection{Baselines}
We compare our \textit{recall-then-verify} system with two state-of-the-art \textit{rerank-then-read} systems.

\noindent
\textbf{R\textsc{efuel}} \cite{DBLP:conf/acl/GaoZNSWNZNAX20} selects 100 top-ranking passages from 1,000 retrieved passages, and predicts answers with a reader based on BART\textsubscript{large} \cite{DBLP:conf/acl/LewisLGGMLSZ20}. It also has a round-trip prediction mechanism, i.e., to generate disambiguated questions based on predicted answers, which are re-fed to the reader to recall more answers.

\noindent
\textbf{JPR} \cite{DBLP:conf/emnlp/MinLCTH21} is a passage reranker which jointly models selected passages. With improved reranking performance, \citet{DBLP:conf/emnlp/MinLCTH21} selected only 10 passages from 100 retrieved passages, and used a reader based on T5-3b which is much larger and more powerful than R\textsc{efuel}'s reader, while requiring no more memory resources than R\textsc{efuel}.

\subsection{Implementation Details}
Our retrieval corpus is the English Wikipedia from 12/20/2018. We finetuned the dense retriever from \cite{DBLP:conf/iclr/IzacardG21} on each multi-answer dataset.
The answer recaller and the answer verifier were initialized with T5-3b; both were pre-trained on NQ and then finetuned on each multi-answer dataset. $\alpha_{neg}$ was 0.1 when finetuning the recaller.
We retrieved 100 passages for a question, and verified each candidate with $k$=10 passages. The threshold $\tau$ for verification was tuned on the dev set based on the sum of F1 scores on all questions (F1 (all)) and questions with multiple answers (F1 (Multi)); the best $\tau$ on W\textsc{eb}QSP/A\textsc{mbig}QA are 0.8/0.5, respectively. Experiments with different model choices for the recaller and different values of $\alpha_{neg}$, $k$ and $\tau$ are shown in Section \ref{sec:ablation_study}.
\text{\small{Please refer to the Appendix for more implementation details.}}

\noindent
\textbf{Memory Constraint: }
\citet{DBLP:conf/emnlp/MinLCTH21} considered a fixed hardware and trained a reader with the maximum number of passages. We follow this memory constraint, under which a reader/verifier based on T5-3b can encode up to 10 passages each of length no longer than 360 tokens at a time.

\subsection{QA Results}
Due to candidate-aware evidence aggregation and a fixed sufficient number of passages distributed to each candidate, our \textit{recall-then-verify} framework can make use of most retrieved supporting passages (see our improvements over OPR in Figure \ref{fig:evid_usage}).
With a higher level of evidence usage, our \textit{recall-then-verify} system outperforms state-of-the-art \textit{rerank-then-read} baselines on both multi-answer datasets, which is shown by Table \ref{tab:multi_ans_res}. Though focused on multi-answer questions, our framework is also applicable to single-answer scenario and achieves state-of-the-art results on NQ. Please refer to the Appendix for more details.
\begingroup
\setlength{\tabcolsep}{3pt} 
\renewcommand{\arraystretch}{1} 
\begin{table}[t!]
    \centering
	\adjustbox{max width=0.4\textwidth}{
    \begin{tabular}{ccccc}
        \hline
		\multirow{2}{*}{System} & \multicolumn{2}{c}{W\textsc{eb}QSP} & \multicolumn{2}{c}{A\textsc{mbig}QA} \\
		\cmidrule(lr){2-3} \cmidrule(lr){4-5}
        & Dev* & Test & Dev & Test \\
        \hline
        R\textsc{efuel} & - & - & 48.3/37.3 & 42.1/33.3 \\
        JPR & 53.6/\textbf{49.5} & 53.1/47.2 & 48.5/37.6 & 43.5/34.2 \\
        \hline
	    Ours & \textbf{55.4}/45.4 & \textbf{55.8}/\textbf{48.8} & \textbf{52.1}/\textbf{41.6} & \textbf{46.2}/\textbf{37.1} \\
	    \hline
    \end{tabular}
    }
    \caption{QA results on multi-answer datasets. The two numbers in each cell are F1 scores on all questions and questions with multiple answers, respectively. \textit{Results on the dev set of W\textsc{eb}QSP can not be directly compared, as we used a different train/dev split\textsuperscript{\ref{footnote:diff_split}}.}}
    \label{tab:multi_ans_res}
\end{table}
\endgroup

\subsection{Ablation Study}
\label{sec:ablation_study}
In this section, we present ablation studies on A\textsc{mbig}QA. Please refer to the Appendix for results on W\textsc{eb}QSP, which lead to similar conclusions.

\subsubsection{Answer Recalling}

\noindent
\textbf{Model Choices for the Answer Recaller}
As shown by Table \ref{tab:ans_recall}, though T5-base is commonly recognized as a much weaker model than T5-3b, a recaller based on T5-base can achieve a high coverage of gold answers, leading to competitive end-to-end performance on the test set.

\noindent
\textbf{Necessity of Verification}
To investigate whether the recaller has the potential to tackle multi-answer questions alone, we tuned the precision of the recaller by varying $\alpha_{neg}$.
As shown in Table \ref{tab:ans_recall}, with increased $\alpha_{neg}$, the recaller learns to recall answers more precisely but still significantly underperforms the overall \textit{recall-then-verify} system.
It is likely that the recaller is trained on false positive passages, which may mislead the recaller to be over-conservative in filtering out hard negative passages.
By contrast, using more evidence for verification is less likely to miss true positive evidence if there is any for a candidate, thus not prone to mislead the verifier.

\noindent
\textbf{Reducing Answer Candidates}
Though only using our recaller for multi-answer QA falls short, the recaller can be trained to shrink down the number of candidates so that the burden on the verifier can be reduced.
As shown by Table \ref{tab:ans_recall}, a small value of $\alpha_{neg}$ helps reduce answer candidates without significantly lowering recall.

\begingroup
\setlength{\tabcolsep}{3pt} 
\renewcommand{\arraystretch}{1} 
\begin{table}[t!]
    \centering
	\adjustbox{max width=.49\textwidth}{ 
    \begin{tabular}{cc|c|ccccc|c}
        \hline
        \multicolumn{2}{c|}{Recaller} & Verifier & \multicolumn{5}{c|}{Dev} & Test \\
        \hline
		T5 & $\alpha_{neg}$ & $\tau$ & $|\mathcal{A}|$ & \# Hit & Recall & Precision & F1 & F1 \\
        \hline
	    3b & 10 & - & 2.2 & 2068/1237 & 54.4/39.0 & 39.6/38.3 & 41.1/34.3 & - \\
        3b & 5 & - & 3.3 & 2206/1328 & 56.8/41.7 & 36.6/36.5 & 39.7/34.7 & - \\
        3b & 1 & - & 7.2 & 2714/1690 & 65.7/50.9 & 22.2/22.7 & 29.7/28.2 & - \\
        3b & 0 & - & 51.2 & \textbf{3364}/\textbf{2211} & \textbf{73.5}/\textbf{61.9} & 3.8/4.6 & 6.8/8.2 & - \\
        \hline
        3b & 0.1 & - & 28.7 & 3288/2141 & 72.6/60.5 & 6.3/7.5 & 10.9/12.7 & - \\
        base & 0.1 & - & 48.4 & 3156/2056 & 70.0/57.9 & 3.3/4.1 & 6.0/7.5 & - \\
        \hline
        \hline
        3b & 0.1 & 0.5 & 28.7 & 2046/1184 & 55.2/37.8 & \textbf{57.7}/\textbf{56.4} & \textbf{52.1}/\textbf{41.6} & \textbf{46.2}/\textbf{37.1} \\
        base & 0.1 & 0.6 & 48.4 & 2051/1181 & 54.8/37.6 & 55.4/54.3 & 50.8/40.8 & 45.8/37.0 \\
	    \hline
    \end{tabular}
    }
    \caption{Performance of recallers on A\textsc{mbig}QA, trained with different models and $\alpha_{neg}$.
    The recaller was paired with a verifier only in the last two rows which 
    show end-to-end QA results.
    \textit{\# Hit} is the number of distinct gold answers verified or recalled, depending on whether the verifier is used or not.}
    \label{tab:ans_recall}
\end{table}
\endgroup

\subsubsection{Answer Verification}
\label{sec:ab_verify}
\begin{figure}[htp]
     \centering
     \begin{subfigure}[b]{0.245\textwidth}
         \centering
         \includegraphics[width=\textwidth]{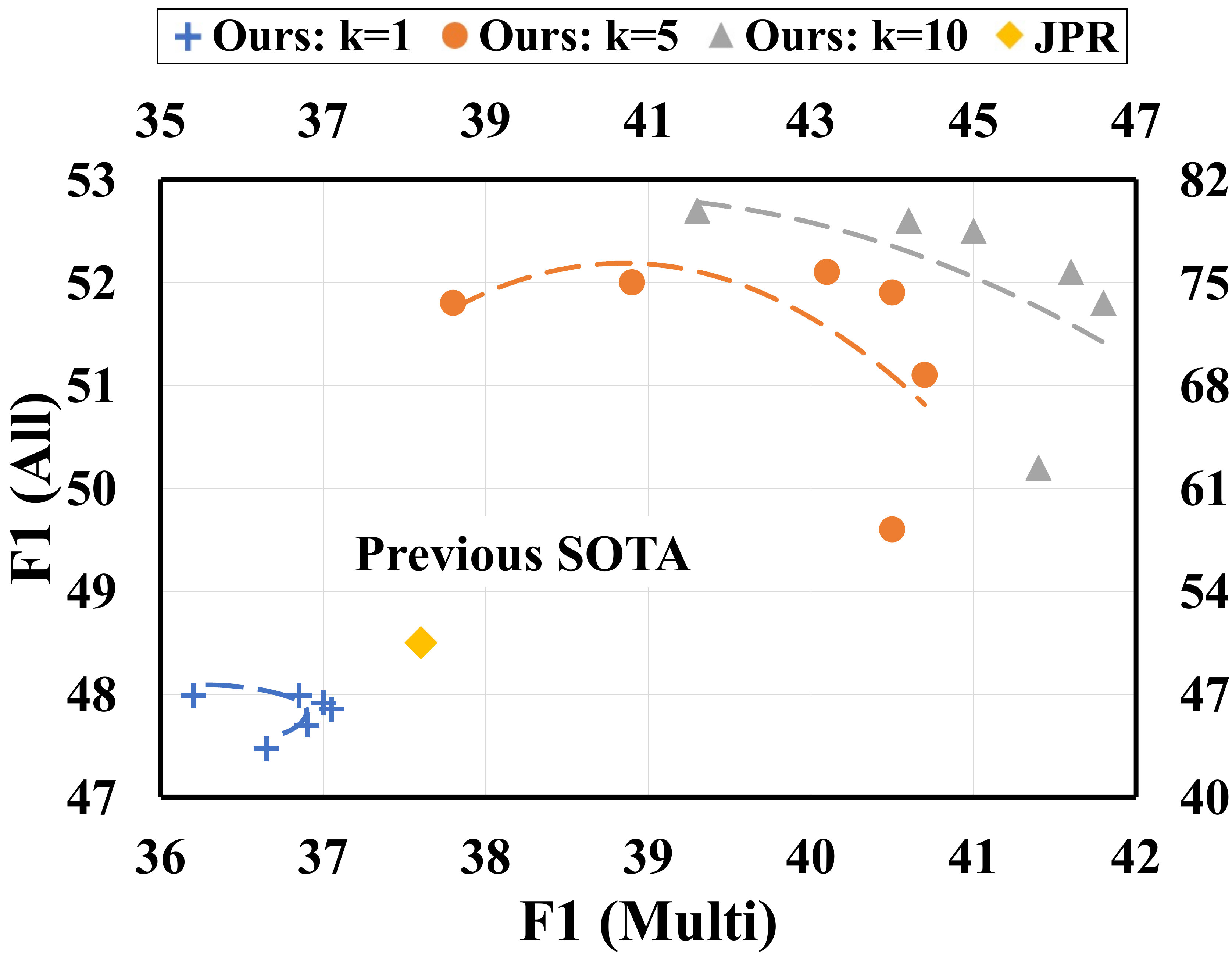}
         \caption{}
         \label{fig:n_ctxs}
     \end{subfigure}
     \hfill
     \begin{subfigure}[b]{0.22\textwidth}
         \centering
         \includegraphics[width=\textwidth]{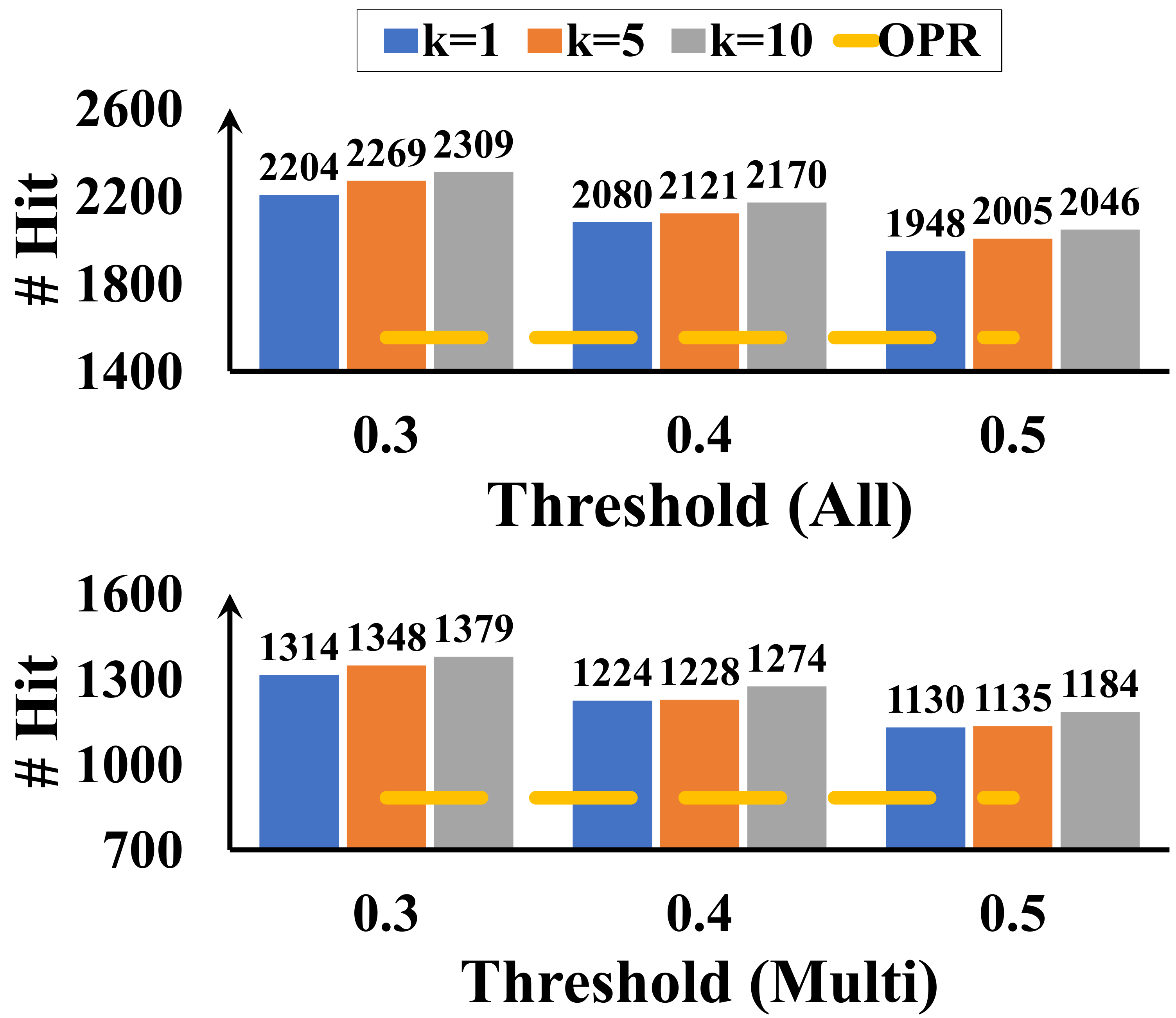}
         \caption{}
         \label{fig:hits}
     \end{subfigure}
    \caption{Performance on the dev set of A\textsc{mbig}QA, with varying $k$ and $\tau$.
    In Figure (a), results with $k$=1 are associated with the top and right axes, while the others are with the bottom and left axes.
    As $\tau$ increases ($\tau \in \{0.3, 0.4, 0.5, 0.6, 0.7, 0.8\}$), points of the same color move from bottom right to top left.
    In Figure (b), \textit{\# Hit} is the number of gold answers with their scores above a threshold.
    \textit{All} and \textit{Multi} denote all questions and questions with multiple answers, respectively.
    }
    \label{fig:k_and_tau}
\end{figure}
\noindent
\textbf{Effect of $k$}
Figure \ref{fig:k_and_tau} shows the benefit of using more evidence for verification. As $k$ increases from 1 to 10, there is a significant boost in F1 scores.

\noindent
\textbf{Effect of $\tau$}
As shown by Figure \ref{fig:n_ctxs}, the balance between recall and precision can be controlled by $\tau$: a lower $\tau$ leads to higher recall and may benefit performance on questions with multiple answers.
With $k$=10, our system outperforms the previous state-of-the-art system for a wide range of $\tau$.
As shown by Figure \ref{fig:hits}, under the best setups ($k$=10, $\tau$=0.5), our system predicts 31.7\% and 34.1\% more gold answers than the system using OPR on all questions and questions with multiple answers, respectively.

\begin{figure}[!t]
    \centering
    \includegraphics[width=0.49\textwidth]{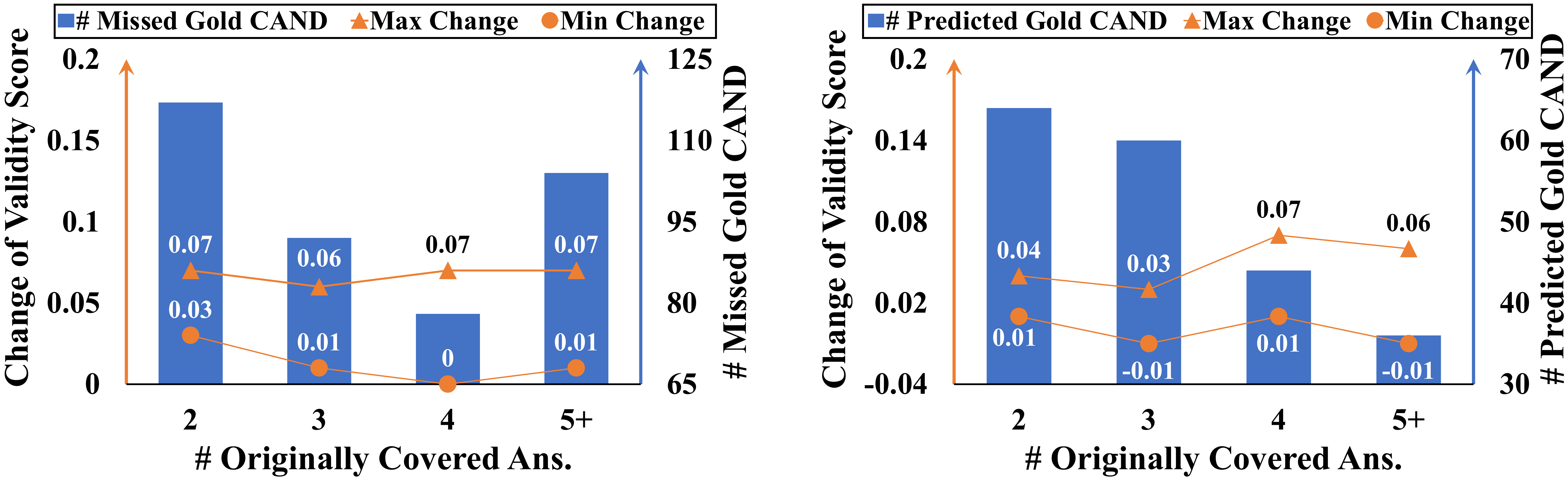}
    \caption{Analysis of how answer verification ($k$=10) is affected by the evidence of other answers on the dev set of A\textsc{mbig}QA. The horizontal axis is the number of answers covered by $\mathcal{E}$. The left axis shows the max and min changes of predicted score of a gold candidate on average after adversarially removing supporting passages of some other answer(s) from $\mathcal{E}$. The left and right graphs are for missed gold candidates (scores $<$ 0.5) and predicted gold candidates (scores $\ge$ 0.5), respectively.
    After attacks, scores of 13.0\% of missed candidates and 3.4\% of predicted ones increased to above and decreased to below 0.5, respectively.
    }
    \label{fig:our_dep}
\end{figure}

\noindent
\textbf{Dependencies among Answers}
Despite being candidate-aware, aggregated evidence $\mathcal{E}$ can also include supporting passages of some other gold answer(s). We therefore investigated how answer verification is affected by the evidence of the other gold answers. Specifically, we attacked the verifier
as follows: a question-candidate pair is a target if and only if 1) the candidate $\hat{a}_i$ is a gold answer and 2) the aggregated evidence $\mathcal{E}_i$ includes at least one supporting passage of some other gold answer(s) that do not cover $\hat{a}_i$; we removed an arbitrary subset of supporting passages of the other gold answer(s) at a time\footnote{Removed passages were replaced with the same number of top-ranking passages that cover no gold answers.} without removing any supporting passages of $\hat{a}_i$, and recorded the worst changes of the predicted validity scores of $\hat{a}_i$.
As shown by Figure \ref{fig:our_dep}, the changes are small, indicating that missed gold candidates with low scores are not mainly suppressed by some other answer(s), and that predicted gold candidates with high scores are verified mainly based on their associated evidence.

\subsection{Error Analysis}
\begingroup
\setlength{\tabcolsep}{3pt} 
\renewcommand{\arraystretch}{1} 
\begin{table}[htb]
    \centering
	\adjustbox{max width=.49\textwidth}{ 
    \begin{tabular}{lr}
        \hline
        \multicolumn{2}{l}{\textbf{Missed Gold Answers}} \\
        Evidence is wrong & 24\% \\
        Evidence is right and straightforward & 76\% \\
	    \hline
	    \multicolumn{2}{l}{\textbf{Wrong Predictions}} \\
	    Predictions are true negatives & 20\% \\
	    Predictions are superficially-different false negatives & 52\% \\
	    Predictions are unannotated false negatives & 28\% \\
	    \hline
    \end{tabular}
    }
    \caption{Analysis of our predictions on the dev set of A\textsc{mbig}QA. Examples are shown in Appendix.}
    \label{tab:manual_analyses_brief}
\end{table}
\begingroup
Among 3,288 recalled gold answers on the dev set of A\textsc{mbig}QA, the answer verifier misses 1,242 of them and outputs 1,323 wrong predictions.
We manually analyzed 50 random samples, 25 of which are missed gold answers and 25 are wrong predictions.
Table \ref{tab:manual_analyses_brief} reports our analysis.

For 76\% of missed gold answers, our evidence aggregator actually aggregates straightforward true positive evidence. Among these missed answers with straightforward evidence, 58\% of them have validity scores higher than 0.2 but lower than the threshold 0.5.
We attacked the verifier on missed gold answers with their validity scores below 0.2 as in Section \ref{sec:ab_verify}, and found that the maximum change of predicted scores on average is small (+0.04), indicating that the low scores can not be attributed to the negative distraction by the other gold answer(s). We conjecture that, as it is difficult even for human annotators to find all valid answers to an open-domain question \cite{DBLP:conf/emnlp/MinMHZ20}, the verifier was trained to refute false negative candidates, resulting in unexpected low scores on some straightforward valid answers.

Notably, 80\% of our ``wrong'' predictions turn out to be false negatives: 52\% of ``wrong'' predictions are semantically equivalent to some annotated answer but are superficially different \cite{DBLP:conf/emnlp/SiZB21}; 28\% of ``wrong'' predictions are unannotated false negatives. Therefore,  it is likely that our system is underrated.

\subsection{Inference Efficiency}
\begingroup
\setlength{\tabcolsep}{3pt} 
\renewcommand{\arraystretch}{1} 
\begin{table*}[htp]
    \centering
	\adjustbox{max width=0.75\textwidth}{
    \begin{tabular}{cc|cccc|c|c|c|c|cc}
        \hline
        \multicolumn{6}{c|}{Answer Recalling} & \multicolumn{2}{c|}{Evidence Aggregation} & \multicolumn{2}{c|}{Answer Verification} & \multicolumn{2}{c}{Overall} \\
        \hline
        T5 & $\alpha_{neg}$ & Sec/Q & $|\mathcal{A}|$ & Recall & Precision & $k$ & Sec/Q & $\tau$ & Sec/Q & Sec/Q & F1 \\
        \hline
        3b & 0.1 & 4.88 & 28.7 & 72.6/60.5 & 6.3/7.5 & 10 & 0.02 & 0.5 & 4.83 & 9.73 & 52.1/41.6 \\
        3b & 0.1 & 4.88 & 28.7 & 72.6/60.5 & 6.3/7.5 & 5 & 0.02 & 0.5 & 2.41 & 7.31 & 51.9/40.5 \\
        base & 0.1 & 0.85 & 48.4 & 70.0/57.9 & 3.3/4.1 & 10 & 0.03 & 0.6 & 8.56 & 9.44 & 50.8/40.8 \\
        base & 0.7 & 0.54 & 16.0 & 63.5/50.1 & 9.5/10.3 & 5 & 0.01 & 0.5 & 1.33 & 1.88 & 50.7/38.2 \\
	    \hline
    \end{tabular}
    }
    \caption{QA performance and inference efficiency of our systems with different configurations on the dev set of A\textsc{mbig}QA. \textit{Sec/Q} denotes seconds per question when using a single V100 GPU. Answer verifiers were all initialized with T5-3b.}
    \label{tab:infer_efficiency}
\end{table*}
\begingroup
\noindent
In this section, we analyze the time complexity of our framework during inference, make comparisons with the state-of-the-art \textit{rerank-then-read} framework JPR, and show how to reduce the computation cost of a \textit{recall-then-verify} system.

For convenience, we denote the encoder length and decoder length as $L_p$ and $L_a$, respectively.

\noindent
\textbf{Recaller vs. Reranker}
The time complexity of answer recalling is $O(|\mathcal{B}| \cdot (L_p^2 + L_a \cdot L_p + L_a^2))$, while that of passage reranking is $O(|\mathcal{B}| \cdot L_p^2 + k \cdot |\mathcal{B}| \cdot L_p + k^2)$. As encoding dominates computation cost (whose time complexity is $O(|\mathcal{B}| \cdot L_p^2)$), given the same model size and $|\mathcal{B}|$, the time complexity of answer recalling and passage reranking is at the same level.

\noindent
\textbf{Verifier vs. Reader}
The time complexity of answer verification is $O(|\mathcal{A}| \cdot (k \cdot L_p^2 + k \cdot L_p))$, while that of the reader is $O(k \cdot L_p^2 + L_a \cdot k \cdot L_p + L_a^2)$. As the reader decodes a sequence of length $L_a$ in an autoregressive way, while the decoding length of the verifier is only 1, the ratio between the inference time of the verifier and that of the reader should be much less than $|\mathcal{A}|$.

\noindent
\textbf{Evidence Aggregator}
Evidence aggregation is significantly faster than answer recalling and verification, as representations of Wikipedia passages are pre-computed. The time complexity is $O(|\mathcal{A}| \cdot (L_p^2 + |\mathcal{B}| \cdot \log k))$ where $L_p^2$ comes from encoding a question-candidate pair with the retriever's question encoder, and $|\mathcal{B}| \cdot \log k$ comes from selecting the top-$k$ relevant passages for a candidate.

One can adjust the computation cost of a \textit{recall-then-verify} system, depending on how much inference efficiency is valued over precision and recall, by (1) choosing a recaller model of proper time complexity\footnote{As shown in Table \ref{tab:ans_recall}, a smaller and faster answer recaller is capable of recalling answers with high coverage.}, (2) tuning $\alpha_{neg}$ to adjust the expected number of candidates $|\mathcal{A}|$ needed for verification, (3) or tuning the number of passages $k$ used for verification.

Table \ref{tab:infer_efficiency} shows QA performance and inference efficiency of our systems with different configurations. Replacing T5-3b with T5-base for the recaller is significantly faster in answer recalling, but is much less precise and produces more answer candidates with the same $\alpha_{neg}$, which increases the burden on the verifier and thus may fail to reduce the overall computation cost if $\alpha_{neg}$ is not raised.
By also increasing $\alpha_{neg}$ and choosing a smaller $k$, as shown by the last row of Table \ref{tab:infer_efficiency}, the overall time needed to answer a question on the dev set of A\textsc{mbig}QA can be reduced to 1.88 sec on a single V100 GPU while also obtaining state-of-the-art F1 scores (50.7/38.2). By contrast, the \textit{rerank-then-read} system from \citet{DBLP:conf/emnlp/MinLCTH21} using a T5-base JPR ($k$=10) and a T5-3b reader is estimated to take 1.51 sec per question\footnote{The average inference time of JPR from \citet{DBLP:conf/emnlp/MinLCTH21} is independent of its parameters given a fixed number of encoded tokens and a fixed decoder length, which can be estimated with a randomly initialized JPR. The average inference time of JPR's reader was estimated with OPR's reader.} with F1 scores of 48.5/37.6.

\section{Conclusion}
In this paper, we empirically analyze the problems of the \textit{rerank-then-read} framework for open-domain multi-answer questions, and propose the \textit{recall-then-verify} framework, which separates the reasoning process of each answer so that 1) we can have a higher level of evidence usage 2) and predicted answers are mainly based on associated evidence and are more robust to distraction by evidence of the other gold answer(s), 3) while also leveraging large models under the same memory constraint.
On two multi-answer datasets, our framework significantly outperforms \textit{rerank-then-read} baselines with new state-of-the-art records.

\section*{Acknowledgements}
This work was supported by the National Science Foundation for Distinguished Young Scholars (with No. 62125604) and the NSFC projects (Key project with No. 61936010 and regular project with No. 61876096). This work was also supported by the Guoqiang Institute of Tsinghua University, with Grant No. 2019GQG1 and 2020GQG0005.

\bibliography{dataset,retriever,diverse_retrieval,passage_reranker,answer_reranker,reader,PLMs,others}
\bibliographystyle{acl_natbib}

\newpage

\appendix

\section{Implementation Details}
\subsection{Retriever}
Our retrieval corpus is the English Wikipedia from 12/20/2018, where articles are split into 100-word passages.
Both OPR and our \textit{recall-then-verify} system share the same passage retriever, which was initialized with the checkpoint released by \cite{DBLP:conf/iclr/IzacardG21} and was finetuned on each multi-answer dataset following DPR \cite{DBLP:conf/emnlp/KarpukhinOMLWEC20}.
Specifically, for each question, we retrieved 100 passages with \citet{DBLP:conf/iclr/IzacardG21}'s checkpoint; for each gold answer $a_i$, we treated top-6 retrieved passages covering $a_i$ as positives, and top-30 retrieved passages covering no gold answer as hard negatives. During finetuning, batch size was set to 128; each question in a batch was paired with one random positive passage and two random hard negatives.

\begingroup
\setlength{\tabcolsep}{3pt} 
\renewcommand{\arraystretch}{1} 
\begin{table}[htp]
    \centering
	\adjustbox{max width=.4\textwidth}{
    \begin{tabular}{cccc}
        \hline
		$|\mathcal{B}|$ & Retriever & W\textsc{eb}QSP (Test) & A\textsc{mbig}QA (Dev) \\
		\hline
		\multirow{2}{*}{5} & DPR\textsuperscript{+} & \textbf{57.0}/\textbf{38.9} & \textbf{55.2}/\textbf{36.3} \\
		& Ours & 56.1/37.7 & 53.2/28.9 \\
		\hline
		\multirow{2}{*}{10} & DPR\textsuperscript{+} & \textbf{59.0}/\textbf{38.6} & 59.3/\textbf{39.6} \\
		& Ours & 57.8/35.9 & \textbf{60.0}/37.7 \\
		\hline
		100 & Ours & 68.0/47.8 & 73.6/57.6 \\
	    \hline
    \end{tabular}
    }
    \caption{Retrieval results in terms of MR\textsc{ecall}. DPR\textsuperscript{+} is the retriever of JPR \cite{DBLP:conf/emnlp/MinLCTH21}. We only report results on the test set of W\textsc{eb}QSP and the dev set of A\textsc{mbig}QA because \citet{DBLP:conf/emnlp/MinLCTH21} used a different train/dev split on W\textsc{eb}QSP and the test set of A\textsc{mbig}QA is hidden.}
    \label{tab:retriever}
\end{table}
\endgroup

Table \ref{tab:retriever} shows the performance of our retriever.
Our retriever underperforms DPR\textsuperscript{+}, the retriever of JPR \cite{DBLP:conf/emnlp/MinLCTH21}, in terms of MR\textsc{ecall}@5 and MR\textsc{ecall}@10.
As DPR\textsuperscript{+} has not been released, it is unknown whether DPR\textsuperscript{+} still covers more gold answers than our retriever when retrieving 100 passages.

\subsection{Answer Recaller \& Answer Verifier}
Our answer recallers used an encoder length of 240 and a decoder length of 40; they were first pre-trained on NQ for 10 epochs and then finetuned on W\textsc{eb}QSP/A\textsc{mbig}QA for 80/20 epochs with early stopping. Batch size was set to 320.
We trained the recallers to decode gold answers covered by a given positive passage (following the order they appear in the passage) and output the ``irrelevant'' token given a negative passage.
Our best system adopts the recaller trained with $\alpha_{neg}$=0.1 because compared with $\alpha_{neg}$=0, the recaller trained with $\alpha_{neg}$=0.1 shrinks down nearly half of answer candidates without a significant drop in recall.

Our answer verifiers used an encoder length of 280; they were first pre-trained on NQ for 3 epochs and then finetuned on W\textsc{eb}QSP/A\textsc{mbig}QA for 30/10 epochs with early stopping. Batch size was set to 320 for $k$=1 and set to 64 for $k\in \{5, 10\}$. The number of invalid answers used for training was set to 10 times the number of valid answers. The best threshold $\tau$ was chosen from $\{0.3, 0.4, 0.5, 0.6, 0.7, 0.8, 0.9\}$ based on F1 scores on the dev set.

We used a flat learning rate of 1e-5 with 500 warm-up steps. All experiments were conducted on a single machine with eight V100 GPUs.

\section{Experiments}
\subsection{Single-Answer QA Result}
\begingroup
\setlength{\tabcolsep}{3pt} 
\renewcommand{\arraystretch}{1} 
\begin{table}[htp]
    \centering
	\adjustbox{max width=0.4\textwidth}{
    \begin{tabular}{ccccc}
        \hline
		System & $k$ & T5 & Dev & Test \\
        \hline
        \cite{DBLP:conf/iclr/IzacardG21} & 100 & large & 51.9 & 53.7 \\
        JPR & 10 & 3b & 50.4 & 54.5 \\
        \hline
	    Ours & 10 & 3b & \textbf{52.8} & \textbf{54.8} \\
	    \hline
    \end{tabular}
    }
    \caption{Exact match scores of different systems on the single-answer dataset NQ. The column \textit{T5} shows the size of the readers of \textit{rerank-then-read} systems and the size of the verifier of our \textit{recall-then-verify} system. \citet{DBLP:conf/iclr/IzacardG21} adopted the \textit{rerank-then-read} framework and used significantly more memory resources for training than JPR and our system.}
    \label{tab:single_ans_res}
\end{table}
\begingroup
\noindent
Though our framework focuses on multi-answer questions, we also experimented on NQ to demonstrate that our framework is applicable to single-answer scenario without suffering from low precision. Specifically, for each question, we only output the candidate with the highest validity score. As shown by Table \ref{tab:single_ans_res}, we slightly outperform previous state-of-the-art \textit{rerank-then-read} systems.

\subsection{Ablation Study on W\textsc{eb}QSP}
\subsubsection{Answer Recalling}
\begingroup
\setlength{\tabcolsep}{3pt} 
\renewcommand{\arraystretch}{1} 
\begin{table}[htp]
    \centering
	\adjustbox{max width=.43\textwidth}{
    \begin{tabular}{cc|c|ccccc|c}
        \hline
        \multicolumn{2}{c|}{Recaller} & Verifier & \multicolumn{5}{c|}{Dev} & Test \\
        \hline
		T5 & $\alpha_{neg}$ & $\tau$ & $|\mathcal{A}|$ & \# Hit & Recall & Precision & F1 & F1 \\
        \hline
        3b & 10 & - & 4.6 & 446/337 & 62.8/51.3 & 44.3/45.9 & 46.2/41.5 & -\\
        3b & 5 & - & 4.9 & 452/335 & 64.4/51.3 & 44.7/44.5 & 46.5/39.9 & -\\
        3b & 1 & - & 7.2 & 495/379 & 67.7/55.9 & 33.5/37.7 & 38.6/38.3 & -\\
        3b & 0 & - & 44.4 & \textbf{669}/\textbf{542} & \textbf{72.5}/\textbf{64.2} & 8.3/11.4 & 12.5/16.5 & -\\
        \hline
        3b & 0.1 & - & 23.9 & 600/476 & 70.7/60.9 & 13.5/17.6 & 18.8/22.4 & -\\
        base & 0.3 & - & 28.9 & 582/460 & 70.8/59.2 & 9.5/13.6 & 14.2/18.3 & -\\
        base & 0.1 & - & 40.7 & 614/494 & 70.9/61.3 & 7.1/10.8 & 11.0/15.3 & - \\
        \hline
	    \hline
	    3b & 0.1 & 0.8 & 23.9 & 414/309 & 58.7/44.6 & \textbf{61.0}/61.0 & \textbf{55.4}/45.4 & \textbf{55.8}/\textbf{48.8} \\
	    base & 0.1 & 0.8 & 40.7 & 425/326 & 57.9/45.6 & 59.9/\textbf{63.5} & 54.2/\textbf{46.9} & 54.5/48.4 \\
	    \hline
    \end{tabular}
    }
    \caption{Performance of recallers on W\textsc{eb}QSP, trained with different models and $\alpha_{neg}$.
    The recaller was paired with a verifier only in the last two rows which show end-to-end QA results.
    }
    \label{tab:ans_recall_webqsp}
\end{table}
\endgroup
\noindent
Table \ref{tab:ans_recall_webqsp} shows the results of recallers on W\textsc{eb}QSP, which were trained with different models and $\alpha_{neg}$. In summary, a weak model suffices to recall answers with high coverage. Using a large and strong model for the recaller benefits precision, but it is still difficult for the recaller alone to answer open-domain multi-answer questions, which necessitates answer verification based on more associated evidence. However, an answer recaller can help reduce the burden on the answer verifier by conservatively filtering out negative candidates.

Though a \textit{recall-then-verify} system with a recaller based on T5-base significantly outperforms JPR on W\textsc{eb}QSP, it lags behind the system with a T5-3b recaller on F1 (all) on the test set. We conjecture that this is because with the same $\alpha_{neg}$=0.1, a T5-base recaller obtains lower recall (69.5/61.4) than a T5-3b recaller (72.1/63.3); a T5-base recaller may need an even lower value of $\alpha_{neg}$ to obtain a higher coverage of gold answers.

\subsubsection{Answer Verification}
\begin{figure}[htp]
    \centering
    \includegraphics[width=0.49\textwidth]{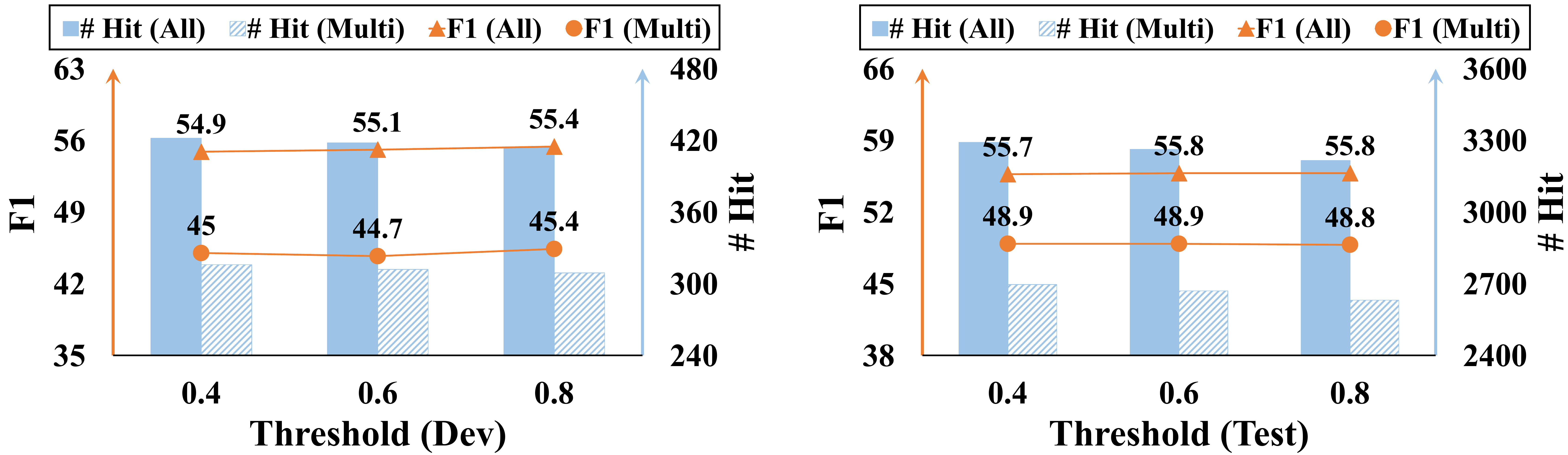}
    \caption{Performance of answer verification ($k$=10) on W\textsc{eb}QSP. \textit{\# Hit} is the number of gold answers with their scores above a threshold.}
    \label{fig:hits_webqsp}
\end{figure}
As shown by Figure \ref{fig:hits_webqsp}, F1 scores on W\textsc{eb}QSP are insensitive to a wide range of $\tau$, while a lower $\tau$ is helpful to predict more gold answers.

\section{Error Analysis}
Table \ref{tab:manual_analyses} reports our error analysis on the dev set of A\textsc{mbig}QA.

\begin{table*}[htp]
    \centering
    \begin{tabularx}{\textwidth}{X}
        \hline
        \textbf{Missed Gold Answers > Evidence is wrong (24\%)} \\
        \hline
        \textbf{Question:} Who does brooke davis have a baby with? \\
        \textbf{Gold Answers:} Julian Baker \\
        \textbf{Missed Gold Answer:} Julian Baker\\
        \textbf{Evidence:} Brooke Davis is happier than ever; preparing to marry Julian Baker ... The Scott family are expecting their second child and Haley feels the baby will be a girl ... \\
        \textbf{Explanation:} Evidence is insufficient to infer Brooke Davis has a baby with Julian Baker. \\
        \hline
        \textbf{Missed Gold Answers > Evidence is right and straightforward (76\%)} \\
        \hline
        \textbf{Question:} What's the most points scored in an nba game? \\
        \textbf{Gold Answers:} 370; 153; 162; 100; 186 \\
        \textbf{Missed Gold Answer:} 162 \\
        \textbf{Evidence:} The 1971-72 team holds franchise records in wins (69), most points scored, and largest margin of victory; both of the latter came in the team's 63 point win versus Golden State (162-99). \\
        \hline
        \hline
        \textbf{Wrong Predictions > Predictions are true negatives (20\%)} \\
        \hline
        \textbf{Question:} When did the song lost boy come out? \\
        \textbf{Gold Answers:} February 12, 2015; January 2015; 4 December 2015; May 9, 2016; 2015; 2017; November 17, 2017 \\
        \textbf{Prediction:} 20 December 2011 \\
        \textbf{Evidence:} ``The Lost Boy'' was written by Holden in 2011 ... Holden recorded it and released as a charity single on 20 December 2011 ... \\
        \textbf{Explanation:} ``Lost Boy'' and ``The Lost Boy'' are different songs. \\
        \hline
        \textbf{Wrong Predictions > Predictions are superficially-different false negatives (52\%)} \\
        \hline
        \textbf{Question:} How much sports are there in the winter olympics? \\
        \textbf{Gold Answers:} fifteen; 86; 98; seven; 102 \\
        \textbf{Prediction:} 15 \\
        \textbf{Evidence:} ... the Winter Olympics programme features 15 sports. \\
        \hline
        \textbf{Wrong Predictions > Predictions are unannotated false negatives (28\%)} \\
        \hline
        \textbf{Question:} How much did it cost rio to host the olympics? \\
        \textbf{Gold Answers:} US\$11.6 billion; US\$13,100,000,000 \\
        \textbf{Prediction:} USD 4.6 billion \\
        \textbf{Evidence:} Indirect capital costs were ``not'' included, such as for road ... Rio Olympics' cost of USD 4.6 billion compares with costs of USD 40-44 billion for Beijing 2008 ... \\
        \hline
    \end{tabularx}
    \caption{Analysis of predictions from our answer verifier. We display all annotated forms of gold answers, which are separated with semicolons.}
    \label{tab:manual_analyses}
\end{table*}

\end{document}